\documentclass[preprint, 12pt]{elsarticle}
\usepackage{booktabs}
\usepackage{graphicx}%
\usepackage{ulem}
\usepackage{multirow}%
\usepackage{amsmath,amssymb,amsfonts}%
\usepackage{amsthm}%
\usepackage{mathrsfs}%
\usepackage{subcaption}%
\usepackage{makecell}%
\usepackage{url}
\usepackage{comment}
\usepackage{array}
\usepackage{lineno}%
\usepackage{xcolor}%
\usepackage{pdflscape}

\journal{X}

\begin{document}

\begin{frontmatter}
\title{Heterogeneous LiDAR Early Fusion and Learned Re-Ranking Strategy for Robust Long-Term Place Recognition in Unstructured Environments}
\author[elche]{Judith Vilella-Cantos\corref{cor1}}
\ead{jvilella@umh.es}
\author[elche]{Juan José Cabrera}
\author[elche]{Mónica Ballesta}
\author[elche]{David Valiente}
\author[elche]{Luis Payá}
\affiliation[elche]{organization={University Institute for Engineering Research, Miguel Hernández University},
        addressline={Avda. de la Universidad s/n, Edificio Innova}, 
            city={Elche},
            postcode={03202}, 
            state={Alicante},
            country={Spain}}
\cortext[cor1]{Corresponding author}

\begin{abstract}
    Robust localization in unstructured environments, such as agricultural fields, is a critical challenge for autonomous systems. LiDAR sensors provide detailed 3D information about the environment and are invariant to lighting conditions. For this reason, LiDAR-based place recognition methods have gained significant attention. In this paper, we propose MinkUNeXt-VINE++, a novel approach that combines early fusion of heterogeneous LiDAR data from two sensors (Livox Mid-360 and Velodyne VLP-16) and a learned re-ranking strategy in inference time. This fusion leverages the strengths of each sensor to provide a more comprehensive representation of the environment. Additionally, the re-ranking approach is particularly important in repetitive environments, such as vineyards, as finding true positives is a major challenge. We evaluated our approach using the TEMPO-VINE dataset, which provides heterogeneous LiDAR data in vineyard environments across different phenological stages. Our results demonstrate that MinkUNeXt-VINE++ significantly improves place recognition performance compared to single-sensor approaches and state-of-the-art methods. MinkUNeXt-VINE++ achieves a 20\% improvement in the Recall@1 metric compared to single-sensor approaches, and +30\% including re-ranking. The code of our method is publicly available for reproduction\footnote{https://github.com/JudithV/MinkUNeXt-VINE\_plusplus}.
\end{abstract}

\end{frontmatter}

\section{Introduction}
Place recognition is a fundamental task in robotics and autonomous systems, enabling them to recognize previously visited locations and localize themselves within an environment. This capability is crucial for various applications, including navigation, mapping, and long-term autonomy. In unstructured environments, such as agricultural fields, place recognition becomes particularly challenging due to the lack of distinctive geometric features, the presence of noise and occlusions, and the dynamic nature of the environment caused by factors such as vegetation growth and weather variations \cite{ou2023place, vilella2026advanced}.

In recent years, Light Detection and Ranging (LiDAR) sensors have gained significant attention for place recognition tasks due to their ability to provide accurate and detailed 3D information about the environment, also being invariant to lighting conditions. Recent works approach the LiDAR Place Recognition (LPR) problem by using Convolutional Neural Networks (CNNs) or transformers. However, in agricultural environments, low-cost approaches are crucial in order to enable the widespread adoption of autonomous systems. State-of-the-art (SOTA)  methods, such as MinkUNeXt-VINE \cite{vilella2026low}, STD \cite{yuan2023std} and SDFT \cite{umemura2024sdft}, focuses on finding robust results with low-cost 3D sensors, such as Livox.

Heterogeneous LiDAR data is becoming increasingly common in real-world applications, as different sensors may be used for different purposes or may be available at different times. For instance, the HeliPR dataset \cite{jung2024helipr} captures data from four different LiDAR sensors. The GEODE dataset \cite{chen2026heterogeneous} also provides heterogeneous LiDAR data captured in degenerated environments. However, most heterogeneous LiDAR datasets are recorded in urban environments and capture very short trajectories, which limits their applicability to place recognition.

In the context of agricultural applications, the TEMPO-VINE dataset \cite{martini2025tempo} offers a unique opportunity to explore the fusion of heterogeneous LiDAR data for place recognition in unstructured environments. This dataset comprises data from both Livox Mid-360 and Velodyne VLP-16 sensors. We use this dataset to evaluate our MinkUNeXt-VINE++, which combines an early fusion strategy for heterogeneous LiDAR data and a learned re-ranking approach to enhance the final ranking of candidate places. The fusion strategy leverages the strengths of each sensor, providing a more comprehensive representation of the environment, while the re-ranking approach refines the results of the initial retrieval stage, improving the performance of place recognition in challenging environments.

The contributions of this work are as follow:
\begin{itemize}
    \item MinkUNeXt-VINE++ is the first early fusion method that combines data from two different LiDAR sensors with different scanning patterns, a Livox Mid-360 and a Velodyne VLP-16, for place recognition in unstructured environments. This fusion strategy leverages the strengths of each sensor, providing a more comprehensive representation of the environment. To the best of our knowledge, there are no existing methods that specifically address the fusion of heterogeneous LiDAR data in a unified representation for place recognition in unstructured environments.
    \item MinkUNeXt-VINE++ includes a learned re-ranking approach that enhances the final ranking of candidate places, taking only two descriptors as an input and with minimal computational overload. The addition of re-ranking presents +30\% of Recall@1 in long-term scenarios.
    \item Extensive evaluation on the TEMPO-VINE dataset \cite{martini2025tempo}, which is the only dataset that provides heterogeneous LiDAR data in vineyard environments. Quantitative results demonstrate that our method improves by 64\% and 48\% the Recall@1 metric with respect with single sensor approaches with Livox Mid-360 and Velodyne VLP-16, respectively.
    \item Comparison of the results of applying our early fused data with our proposed preprocessing technique to SOTA methods. The results demonstrate that our preprocessing techniques improve urban-tailored LPR methods as well.
\end{itemize}

This paper in organized as follows: Section \ref{sec:rw} provides a comprehensive overview of the LiDAR place recognition problem, with special emphasis in fusion approaches and re-ranking contributions. Section \ref{sec:methodology} describes the architecture of our proposed MinkUNeXt-VINE++ method, detailing the fusion strategy and the learned re-ranking approach. Section \ref{sec:experimental_setup} describes the dataset used for our experiments, the evaluation metrics, and the training and test sets. Section \ref{sec:ablation_study} presents an ablation study justifying the design decisions of our method. This study analyze the effects of the fusion strategy and learned re-ranking on its performance in cross-sensor and cross-season scenarios. It also highlights the impact of each contribution on other pipelines and datasets, such as the Bacchus Long-Term (BLT) dataset \cite{polvara2024bacchus}. Section \ref{sec:discussion} presents the results of our proposed method for every campaign in the TEMPO-VINE dataset and compares them with the results from other SOTA LPR solutions. Finally, Section \ref{sec:conclusions} concludes the paper and outlines future research directions.

\section{Related Works}\label{sec:rw}
\subsection{LiDAR Place Recognition}\label{subsec:rw_lpr}
LiDAR sensors provide accurate and detailed 3D information about the environment, making them ideal for place recognition tasks. LiDAR-based place recognition methods can be broadly categorized into two main approaches: geometry-based methods and learning-based methods. Geometry-based methods typically extract handcrafted features from LiDAR point clouds, such as geometric descriptors or keypoint-based features, to represent the environment. The most representative example of this type of methods is Scan Context \cite{kim2018scan}, which projects the point cloud to its Bird's Eye View (BEV) representation and, dividing the spaces in sectors, selects the highest height value of the points that fall into the sector. Learning-based methods, on the other hand, leverage deep learning techniques to learn discriminative features directly from raw LiDAR data. The most representative example of a learned LPR method is PointNetVLAD \cite{uy2018pointnetvlad}, which uses a PointNet architecture to extract features from the point cloud and a VLAD layer to aggregate these features into a global descriptor for place recognition. Both geometry-based and learning-based methods have shown promising results in various environments, but they often struggle in unstructured environments due to the lack of distinctive geometric features and the presence of noise and occlusions.

\subsection{Fusion techniques}\label{subsec:rw_fusion}
Over the last few years, there has been a significative increment in the use of fusion techniques in the field of LPR. These techniques have demonstrated their effectiveness in improving the performance of LPR algorithms by combining complementary information from multiple sources. There exists three types of fusion techniques: early fusion, late fusion, and mid fusion. Early fusion methods combine raw data from multiple sensors before feature extraction, allowing the model to learn joint representations \cite{bernreiter2021spherical}. On the other hand, late fusion methods (p. e., MinkLoc++ \cite{komorowski2021minkloc++}) extract features independently from each sensor and then combine these features during inference time. Mid fusion methods combine the features of each modality in intermediate layers of the network \cite{pan2024camera}.

Much of the current literature aims to fuse LiDAR and camera data because the latter helps capture fine-grained details and textures often missing from LiDAR data. For example, in LCPR \cite{zhou2023lcpr}, a multimodal place recognition descriptor is obtained by a neural network that processes multi-view panoramic images and a range image from LiDAR with multi-scale attention mechanisms. This procedure has been shown to produce a robust, yaw-invariant descriptor in an urban context. Another example is the innovative quantum-inspired approach of Zhang et al. \cite{zhang2024quantum}. This late-fusion approach combines LiDAR and camera data by encoding them into quantum states and applying a convolutional neural network to create descriptors from both modalities. On a different note, PRFusion \cite{wang2024prfusion} is a multi-modal place recognition method that fuses RGB images and LiDAR data. It introduces a novel fusion block that enhances the interaction between the features of both modalities. The fusion block combines global and local features from the RGB images and LiDAR data. If the extrinsic transformation between sensors is available, the block uses the latter. The block also uses a neural diffusion module based on Beltrami flow \cite{chamberlain2021beltrami}. This allows the model to learn more discriminative features for place recognition.

However, although methods such as HeLiOS \cite{jung2025helios} process a single point cloud independently using a shared encoder with spherical transformers, the literature has less explored the explicit fusion of data from different LiDAR sensors. To the best of our knowledge, there are no existing methods that specifically address the fusion of heterogeneous LiDAR data in a unified representation for place recognition in unstructured environments, which is the focus of our work.

\subsection{Re-ranking strategies}\label{subsec:rw_reranking}
Re-ranking strategies have been widely used in information retrieval tasks to improve the final ranking of retrieved candidates. In the context of LPR, several approaches exist to refine the initial ranking of candidate places. For example, HOTFLoc++ \cite{griffiths2025hotfloc++} combines multi-scale geometric verification with learned feature matching to assess geometric consistency with the query at multiple feature granularities. However, performing multi-scale geometric verification across multiple granularities introduces a heavy computational burden and high latency during inference. The VGGT-MPR framework \cite{xu2026vggt} introduces a training-free re-ranking mechanism for multimodal place recognition by leveraging the VGGT foundation model for robust cross-view keypoint tracking. Rather than relying on costly geometric alignments, it refines retrieval candidates using confidence-based correspondence scores extracted from semantically static regions, entirely bypassing the need for domain-specific retraining. Nevertheless, the reliance on a large foundation model and continuous keypoint tracking remains highly memory-intensive and computationally demanding for real-time deployment.

Other re-ranking strategies rely primarily on transformers. For instance, TReR \cite{barros2023trer} is a lightweight transformer-based re-ranking approach that takes as input the top K candidates retrieved by the main LPR pipeline and learns to re-rank them based on their embeddings. On the other hand, ETR \cite{zhang2023etr} also presents a transformer-based re-ranking approach in the context of visual place recognition. However, ETR takes as an input a combination of the query and candidate descriptors, allowing it to learn more complex relationships between the query and candidates. Both TReR and ETR have shown significant improvements in LPR performance, demonstrating the importance of re-ranking strategies in refining the results of initial retrieval stages. Nevertheless, their reliance in transformers is computationally expensive, and it may not be suitable for real-time applications or resource-constrained environments. 

\begin{figure}[h]
    \centering
    \includegraphics[width=0.85\textwidth]{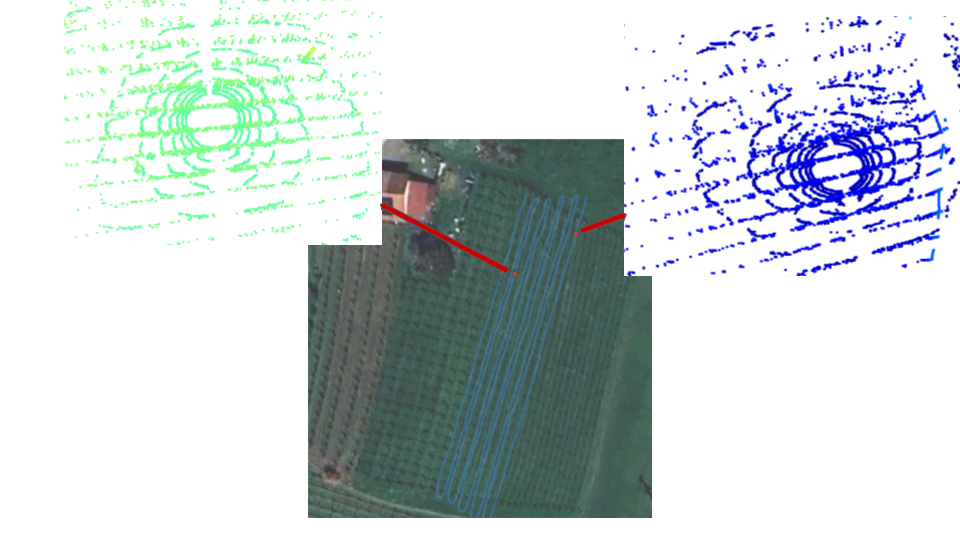}
    \caption{Point cloud similarity between two different places in the same vineyard. While the point cloud in the left corresponds to an intra-row place, the point cloud in the right is more close to a turnpoint section. Furthermore, both point clouds pertain to distant rows of the vineyard.}
    \label{fig:similarity}
\end{figure}

Re-ranking strategies are particularly important in agricultural environments because the high similarity between places and crop occlusions in the latter stages can lead to a high number of false positives in the initial retrieval stage. Figure \ref{fig:similarity} shows an example of point cloud similarity between two distant places in the same vineyard. As shown in the figure, two places within the same trajectory that are in very different positions (intra-row and extreme) show little difference. This makes retrieval a non-trivial problem in these types of settings. By refining the ranking of candidates, re-ranking approaches can significantly improve the performance of LPR algorithms in these challenging environments.

\section{Methodology}\label{sec:methodology}
\begin{figure}[h]
    \centering
    \includegraphics[width=\textwidth]{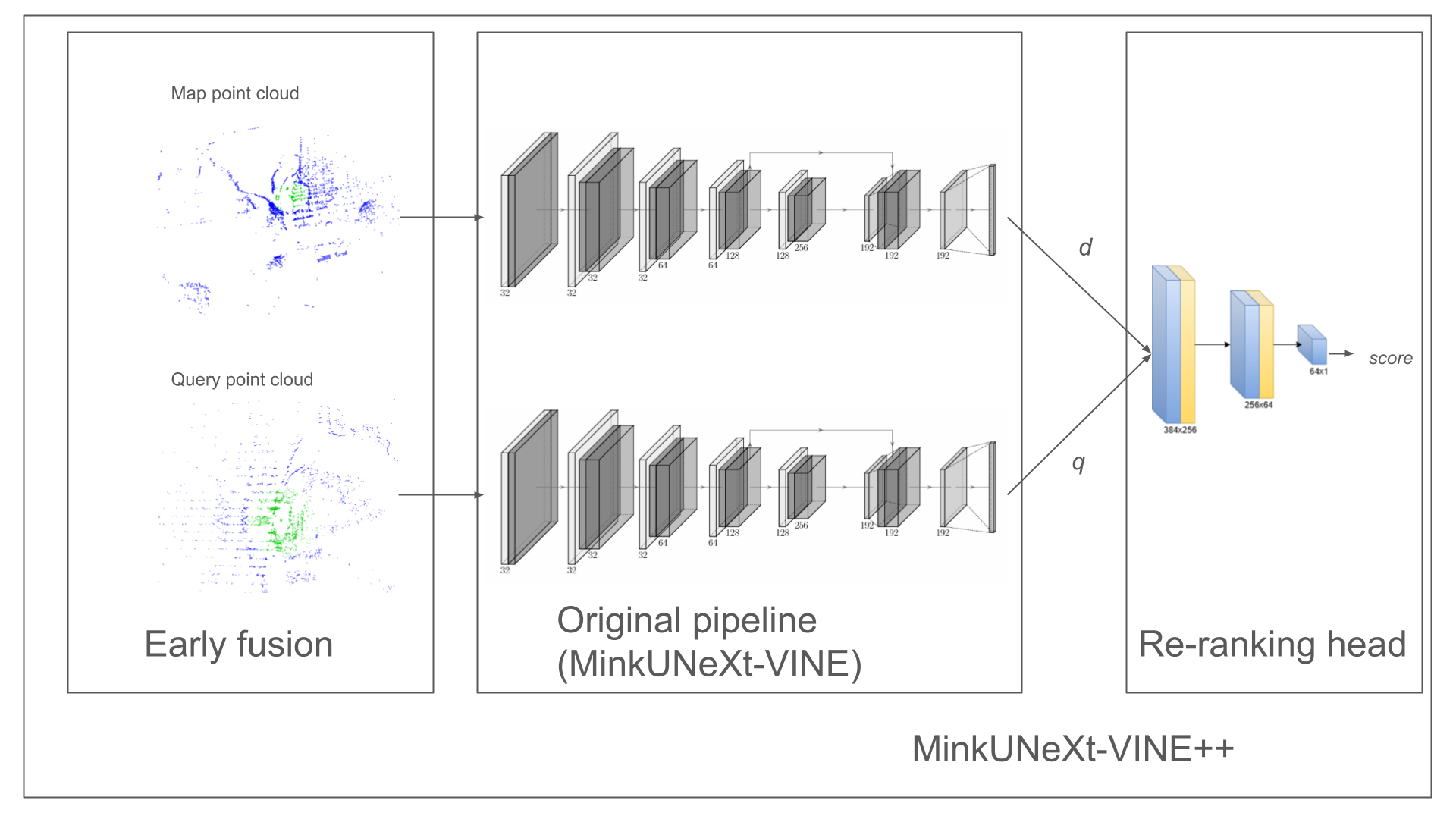}
    \caption{Architecture of our proposed MinkUNeXt-VINE++ method. Firstly, two point clouds from Livox and Velodyne sensors are fused following our proposed fusion strategy. Then, the fused point cloud is processed by the main pipeline, which is based on MinkUNeXt-VINE \cite{vilella2026low}. Finally, a learned re-ranking head takes as input the embeddings of a query and a database point cloud retrieved by the main pipeline and learns to reorder them based on a resulting similarity score.}
    \label{fig:methodology}
\end{figure}

On top of the vineyard-tailored LPR method MinkUNeXt-VINE \cite{vilella2026low}, we propose two main contributions to enhance the performance of LPR in unstructured environments: a heterogeneous LiDAR early fusion strategy and a post-training re-ranking approach. The fusion strategy allows us to leverage the strengths of both Livox Mid-360 and Velodyne VLP-16 sensors provided by the TEMPO-VINE dataset \cite{martini2025tempo}, while the re-ranking addition enhances the final retrieval results. These two contributions lead us to our proposed MinkUNeXt-VINE++ method. Figure \ref{fig:methodology} illustrates the overall architecture of our proposed method, highlighting the fusion and re-ranking components.

\subsection{Sensor fusion}
We consider the main differences between two LiDAR sensors with different scanning patterns when fusing data from them. The Livox Mid-360 is a solid-state LiDAR sensor that provides a wide field of view (FoV) of 360 degrees horizontally and 38.4 degrees vertically. Its non-repetitive scanning pattern allows it to capture more points in a given area than traditional spinning LiDARs. In contrast, the Velodyne VLP-16 is a spinning, mechanical LiDAR sensor that provides a narrower FoV of 360 degrees horizontally and 30 degrees vertically. It uses a repetitive scanning pattern, resulting in a more uniform distribution of points, though it may miss certain areas due to occlusions or limited resolution.

According to Betterncourt et al. \cite{bettencourt2025comparison}, Livox provides reliable information at short ranges, while Velodyne performs better at long ranges. Therefore, we propose a fusion strategy that leverages the strengths of each sensor. 

First, we downsample the point clouds from both sensors to a common resolution. For coherence, we then transform the points of the Velodyne sensor to the common frame between both sensors, using the extrinsic transformation. Then, we define a distance threshold that distinguishes between short- and long-range areas. Specifically, we chose 10 meters as the threshold for the short-range area. We select points from the Livox source exclusively for the short-range area and from the Velodyne source exclusively for the long-range area. This method yields a fused point cloud that combines the strengths of both sensors without the need for complex approaches or high computational overhead. Finally, we combine the selected points from both sensors to create a fused point cloud for LPR. 

\begin{figure}[h]
    \centering
    \includegraphics[width=\textwidth]{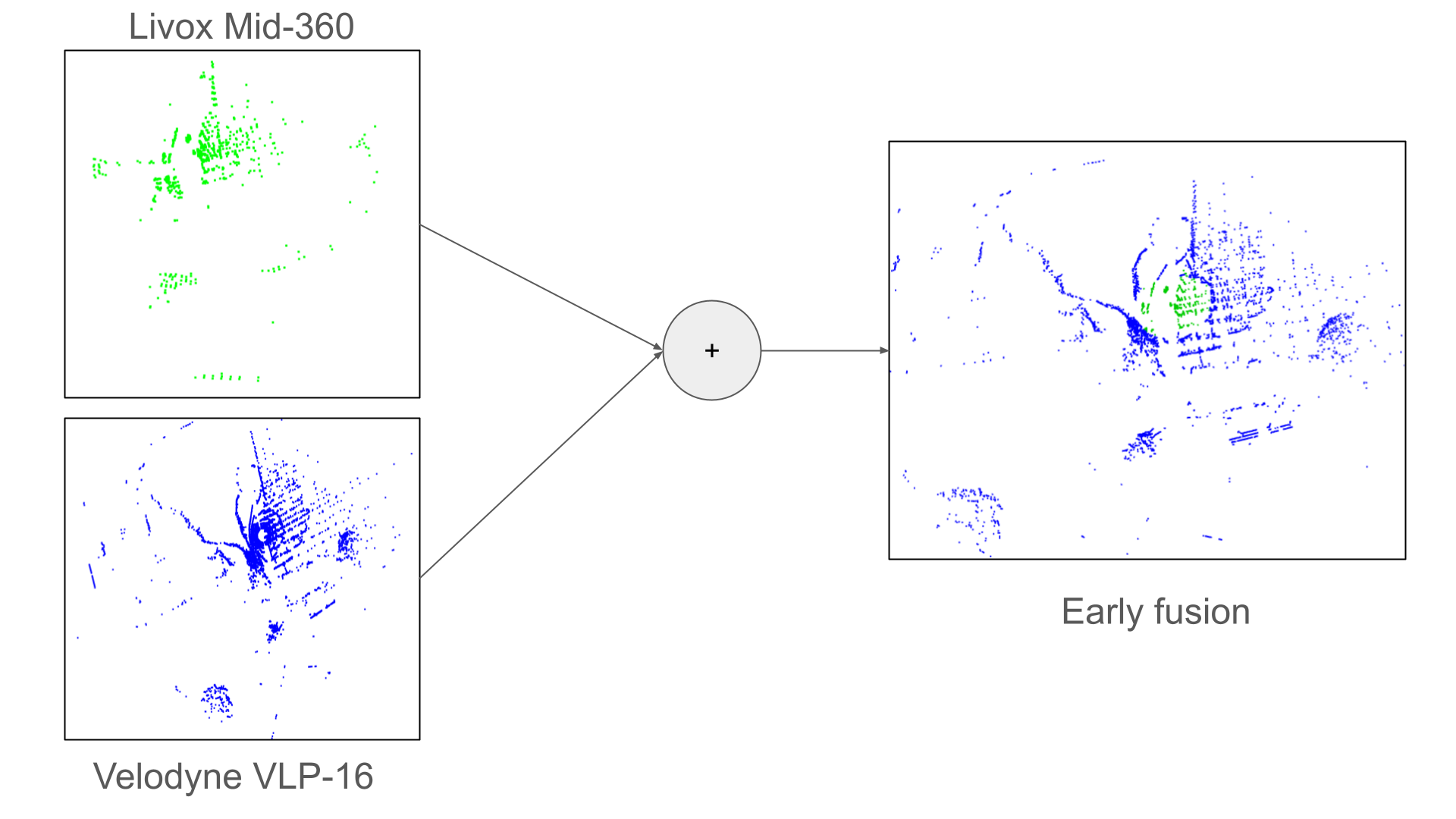}
    \caption{Example of a fused point cloud from the February campaign following our proposed fusion strategy. The points in green correspond to the Livox Mid-360 sensor, while the points in blue correspond to the Velodyne VLP-16 sensor.}
    \label{fig:fusion}
\end{figure}

Figure \ref{fig:fusion} illustrates a fused point cloud from the February campaign (which had better visibility between the rows) using this strategy. In addition to using the closest points from the Livox sensor, we can see that the points from the Velodyne sensor provide a more uniform distribution, which is particularly beneficial for the farthest areas of the vineyard where the Livox sensor may have difficulty capturing sufficient detail. Additionally, the final point cloud is uniformly downsampled, enabling us to maintain consistent point density across the entire cloud, regardless of the original density of each sensor. This is important for LPR methods because it ensures the model receives a balanced representation of the environment.

\subsection{Learned re-ranking}\label{subsec:learned_reranking}
\begin{figure}[h]
    \centering
    \includegraphics[width=0.35\textwidth]{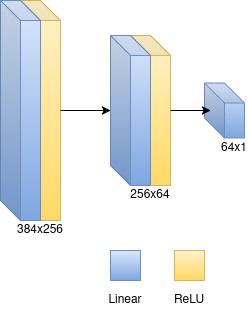}
    \caption{Architecture of the learned re-ranking additional head in inference.}
    \label{fig:reranking}
\end{figure}

Once the training of the main pipeline is done, we introduce a learned re-ranking head that takes the embeddings of the top K candidates retrieved by the main pipeline and learns to reorder them. To do so, we introduce a neural network architecture in the style of a Multi-Layer Perceptron (MLP). This head is comprised by three linear layers. After each of them, we apply a Rectified Linear Unit (ReLU) activation function for non-linearity. This structure has an input dimensionality of 384 because the input is formed by concatenating two 192-dimensional descriptors: a query descriptor $q$ and a descriptor from the database $d$, created after the completion of the training of the main pipeline (MinkUNeXt-VINE \cite{vilella2026low}). Figure \ref{fig:reranking} shows the architecture of the re-ranking head.

The re-ranking head is trained using a Binary Cross Entropy (BCE) loss \cite{baum1987supervised}, where the positive samples are the true matches and the negative samples are the false matches among the top $K$ candidates. The formula of the BCE loss function is as follows:

\begin{equation}
    \mathcal{L} = -\frac{1}{N} \sum_{i=1}^{N} [y_i \log(p_i) + (1 - y_i) \log(1 - p_i)]
    \label{eq:binary_cross_entropy}
\end{equation}
where $N$ is the number of samples, $y_i$ is the true label (1 for positive matches and 0 for negative matches), and $p_i$ is the predicted probability of being a positive match.

This additional head is trained in inference time for 25 epochs, and the batch size is 1024. The optimizer used is Adam \cite{kingma2014adam} with a learning rate of 0.001. The re-ranking head is designed to be lightweight and efficient, ensuring that it does not introduce significant computational overhead during inference. For large evaluations involving at least three campaigns and their corresponding trajectories, each epoch takes around one minute on an Nvidia A30 GPU.

\section{Experimental Setup}\label{sec:experimental_setup}
\subsection{Dataset}\label{subsec:dataset}
The dataset used for our experiments is the TEMPO-VINE dataset \cite{martini2025tempo}. This is the only dataset that provides heterogeneous LiDAR data in an agricultural environment (vineyards). It consists of data collected from two different LiDAR sensors: a Livox Mid-360, which is a solid-state LiDAR sensor with a wide field of view, and a Velodyne VLP-16, which is a mechanical spinning LiDAR sensor with a narrower field of view. The dataset covers the full phenological cycle of the vineyard, as it was captured following different campaigns from February of 2025 until November of the same year. The dataset is organized into sequences, each corresponding to a specific campaign. Each sequence contains synchronized data from both LiDAR sensors, along with GPS-RTK information that serve as ground truth for localization tasks. There are three different sequences for each campaign, which correspond to three different runs of the vineyard. Each run consists of a trajectory that covers several rows of the vineyard, providing a comprehensive representation of the environment. "run1" covers ten rows of the vineyard recorded in a forward direction. Similarly, "run2" covers eight rows of the vineyard, the last four rows recorded in an inverse direction than "run1". Finally, "run3" covers three rows of the vineyard, performing a close-loop trajectory. Differently from the other two runs, "run3" was introduced in the last campaign of June, being not available for the first campaings. Regarding the pergola environment, it contains two different routes per each campaing, both covering the same trajectory: four rows. Both of the environments were recorded in the same campaigns, following the same phenological stages, and with the same sensors. A summary of the dataset structure (focused on the trellis setting, since this is what we use for our experiments) is provided in Table \ref{tab:dataset_structure}.

\begin{table}[h]
    \centering
    \begin{tabular}{l|c|c|c}
    \hline
     & \textbf{run1} & \textbf{run2} & \textbf{run3} \\
    \hline
    \textbf{Rows captured} & 10 & 8 & 3 \\
    \textbf{Loop-closure}  & $\times$ & $\times$ & \checkmark \\
    \hline
    \end{tabular}
    \caption{Structure of the TEMPO-VINE dataset.}
    \label{tab:dataset_structure}
\end{table}

\begin{figure}
    \centering
    \includegraphics[width=0.9\textwidth]{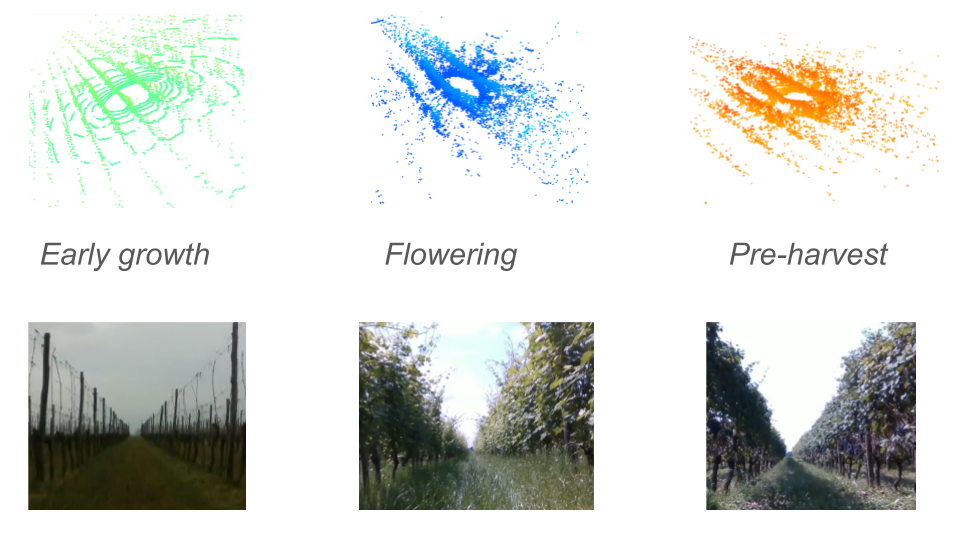}
    \caption{Examples of data captured from a Velodyne VLP-16 and the RGB-D camera from the TEMPO-VINE dataset in different phenological phases. The first column corresponds to the \textit{early growth} stage, the second column corresponds to the \textit{flowering} stage, and the third column corresponds to the \textit{pre-harvest} stage.}
    \label{fig:dataset}
\end{figure}

The dataset provides a challenging testbed for evaluating the performance of LPR algorithms in unstructured environments, as it includes various environmental conditions, such as changes in vegetation growth, lighting conditions, grass growth, and weather variations. Figure \ref{fig:dataset} shows examples of camera images and point clouds in different phenological stages, illustrating the variations in the environment across seasons. The different phenological stages depicted are \textit{early growth}, \textit{flowering}, and \textit{pre-harvest}. The \textit{early growth} phase includes the months of February, March and April. The \textit{flowering} phase includes the months of June (early and late) and early July. The \textit{pre-harvest} phase includes the months of July (late), August and September. 

While the adjacent rows of the vineyard are clearly visible in the \textit{early growth} stage, the vegetation growth severely affects the visibility in the \textit{flowering} and \textit{pre-harvest} stages. The early growth stage is characterized by young plants and limited vegetation, which makes identifying and tracking features easier. In contrast, the \textit{flowering} and \textit{pre-harvest} stages are marked by dense vegetation. Concretely, the \textit{flowering} stage is characterized by the presence of flowers, while the \textit{pre-harvest} stage is characterized by the presence of the developed fruits. The presence of these variations in the dataset allows us to evaluate the robustness and generalization ability of our proposed method across the different stages of the crop.

\subsection{Evaluation metrics}\label{subsec:metrics}
For evaluating the performance of our method, we use the Recall@1\% and Recall@1 metrics. Recall@1\% measures the percentage of queries for which at least one correct match is found within the top 1\% of the retrieved candidates. Recall@1, on the other hand, measures the percentage of queries for which the top retrieved candidate is a correct match. These metrics are commonly used in place recognition tasks to evaluate the effectiveness of retrieval methods. Equation \ref{eq:recall} defines the Recall@K metric, which can be adapted to both Recall@1\% and Recall@1 by setting K accordingly.
\begin{equation}\label{eq:recall}
    \text{Recall@K} = \frac{\text{Queries with at least one correct match in top K}}{\text{Total queries}} \times 100\%
\end{equation}

In our experiments, we consider crucial obtaining a higher value in the Recall@1 metric, as it indicates that the top retrieved candidate is a correct match, which is essential for accurate localization in real-world applications. However, we also report the Recall@1\% metric to provide a more comprehensive evaluation of the retrieval performance, especially in scenarios where multiple candidates may be relevant.

\subsection{Training and test sets}\label{subsec:train_test_split}
Every experiment in this paper is performed using the TEMPO-VINE dataset \cite{martini2025tempo}, while we also report results for the BLT dataset \cite{polvara2024bacchus} for the ablation study of the re-ranking contribution (Section \ref{subsec:re-ranking_ablation}). 

Taking into account the structure of the TEMPO-VINE dataset displayed in Table \ref{tab:dataset_structure}, we use "run1" for training and "run2" for testing. Starting in June, we also include "run3" in the training set. On the other hand, as the BLT dataset contains five rows of a vineyard that is half the size of the TEMPO-VINE dataset and was acquired using a loop-closure trajectory. For this dataset, we use the evaluation protocol proposed by PointNetVLAD \cite{uy2018pointnetvlad}, which consists on using specific zones of the dataset for testing and the rest of the zones for training. As test zones, we designated a GPS position with a radius of 15 meters. Setting a larger radius for the test zones would lead to an imbalance between positive and negative samples, which could affect the training process and the performance of the model.

Regarding the classification of positive and negative samples, we consider two samples to be positive if they are within a 5-meter radius of each other, as determined by the provided GPS-RTK information. Conversely, if the distance between the two samples is greater than 13 meters, they are considered a negative match.  Although the threshold for negative samples is typically set at 25 meters for urban LiDAR datasets, we have chosen more conservative thresholds of 13 meters for the TEMPO-VINE dataset and 12 meters for the BLT dataset. This is because agricultural datasets cover much smaller areas than urban datasets, such as the Oxford RobotCar Dataset \cite{maddern20171}. Setting a larger threshold for negative samples could lead to an imbalance in the number of positive and negative samples, affecting the training process and model performance. Setting a more conservative threshold ensures a more balanced distribution of positive and negative samples, which is crucial for training an effective LPR model in unstructured environments. Finally, in the retrieval phase, a prediction is considered correct if the distance between the two descriptors is 5 meters or less.


\section{Ablation study}\label{sec:ablation_study}
In this section, we justify the design choices of our method through an ablation study. We analyze the effects of the fusion strategy and height-based feature aggregation on the performance of our method in cross-sensor and cross-season scenarios. First, to evaluate the early fusion ablation experiments, we use the three campaigns considered in the \textit{flowering} phase, using "run1" (also "run3" if available) as the training set and "run2" as the test set, as was mentioned under Section \ref{subsec:train_test_split}. Then, for the re-ranking evaluation, we also use the \textit{early growth} and \textit{pre-harvest} bundles, containing three campaigns each, in order to assess the impact of re-ranking on the overall performance across all the phenological stages. Finally, for the cross-season evaluation, we use the same training and test sets, but we train and test our method using different campaigns in order to assess the generalization ability of our proposed contributions, as will be specified in Section \ref{subsec:ablation_cross_season}. Using multiple campaigns for the different stages of the ablation study allows us to evaluate the robustness of our method in different environmental conditions and phenological stages, crucial for long-term autonomy in agricultural settings, which is one of the main objectives of this paper. 

\subsection{Effect of the early fusion strategy}\label{subsec:ablation_fusion}
In order to demonstrate the effectiveness of our proposed fusion strategy, we compare the results of MinkUNeXt-VINE for each of the sensors of the TEMPO-VINE dataset with the results of our preprocessed fused data following the proposed methodology. We also make a comparison with other SOTA urban-tailored LPR methods, such as PointNetVLAD \cite{uy2018pointnetvlad}, MinkLoc3Dv2 \cite{komorowski2022improving} and MinkUNeXt \cite{cabrera2025minkunext}. The results are presented in Table \ref{tab:fused_data_ablation}.

\begin{table}[h]
    \centering
    \begin{tabular}{c|c|cc}
    \hline
    \textbf{Sensor} & \textbf{Method} & \textbf{Recall@1\%} & \textbf{Recall@1}\\
    \hline
    Livox Mid-360 & PointNetVLAD & 33.40 & 5.96\\
    Velodyne VLP-16 & PointNetVLAD & 44.49 & 10.91\\
    Early Fusion & PointNetVLAD & \textbf{49.28} & \textbf{10.99}\\
    \hline
    Livox Mid-360 & MinkLoc3Dv2 & 15.51 & 2.15\\
    Velodyne VLP-16 & MinkLoc3Dv2 & 26.08 & \textbf{11.70}\\
    Early Fusion & MinkLoc3Dv2 & \textbf{29.80} & 3.65\\
    \hline
    Livox Mid-360 & MinkUNeXt & 25.43 & 4.65\\
    Velodyne VLP-16 & MinkUNeXt & 20.59 & 5.47\\
    Early Fusion & MinkUNeXt & \textbf{44.40} & \textbf{11.30}\\
    \hline
    Livox Mid-360 & MinkUNeXt-VINE & 62.79 & 26.85\\
    Velodyne VLP-16 & MinkUNeXt-VINE & 88.38 & 52.80\\
    Early Fusion & MinkUNeXt-VINE & \textbf{96.28} & \textbf{61.41}\\
    \hline
    \end{tabular}
    \caption{Recall@1\% and Recall@1 results comparison before and after our proposed fusion strategy on SOTA urban-tailored LPR methods.}
    \label{tab:fused_data_ablation}
\end{table}

The ablation results presented in Table \ref{tab:fused_data_ablation} validate the efficacy of our proposed sensor fusion strategy across various SOTA LPR architectures. The proposed early fusion approach consistently delivers substantial performance gains over single-sensor baselines. This synergistic effect is most evident in the base method, MinkUNeXt-VINE. When evaluated on the preprocessed fused data, it achieves a remarkable Recall@1\% of 96.28\% and a Recall@1 of 61.41\%, drastically outperforming its single-sensor configurations. Specifically, the fusion strategy boosts Recall@1 by +34.56\% over the standalone Livox Mid-360 sensor and by +8.61\% over the Velodyne VLP-16 sensor. These results highlight the network's superior capacity to exploit the complementary spatial features and varying FoV geometries provided by heterogeneous LiDARs.

Furthermore, the benefits of the early fusion approach generalize robustly across other urban-tailored LPR baselines. For instance, MinkUNeXt experiences a major performance surge in Recall@1\%, increasing by +18.97\% and +23.81\% compared to its respective Livox and Velodyne baselines, while simultaneously doubling its strict Recall@1 accuracy. Similarly, PointNetVLAD shows steady improvements across both metrics, with Recall@1\% increasing by +15.88\% (over Livox) and +4.79\% (over Velodyne). This demonstrates that our data preprocessing methodology enhances overall scene distinctiveness, independent of the underlying neural network backend. However, an interesting edge case occurs with MinkLoc3Dv2. While the early fusion strategy successfully improves its Recall@1\% performance (+3.72\% over the Velodyne baseline), it suffers a performance degradation in the strict Recall@1 metric (dropping from 11.70\% to 3.65\%). This anomaly may be attributed to Livox Mid-360's degradated performance in this specific architecture, which could introduce noise into the fused representation and negatively impact the model's ability to identify the single best match. 

Overall, the empirical evidence confirms that our proposed early fusion methodology effectively mitigates individual sensor limitations, resulting in a significantly more comprehensive and robust environmental representation for place recognition. The next subsection further analyzes the generalization capabilities of our fusion strategy in cross-sensor scenarios, demonstrating its robustness and adaptability to different sensor configurations.

\subsubsection{Cross-sensor evaluation}\label{subsubsec:ablation_cross_sensor}
We evaluate the performance of the early fusion component of the base method MinkUNeXt-VINE in a cross-sensor scenario. Specifically, we train our model using data from one sensor (e.g., Livox Mid-360) and test it using data from the other sensor (e.g., Velodyne VLP-16), and vice versa. Finally, the third scenario consists of training our model using the fused data and testing it using data from each of the sensors separately. Table \ref{tab:cross_sensor} shows the results of the cross-sensor evaluation, also using the \textit{flowering} phase training and test set. 

\begin{table}[h]
    \centering
    \begin{tabular}{c|c|cc}
    \hline
    \textbf{Train Sensor} & \textbf{Test Sensor} & \textbf{Recall@1\%} & \textbf{Recall@1}\\
    \hline
    Livox Mid-360 & Velodyne VLP-16 & 60.00 & 18.30\\
    Velodyne VLP-16 & Livox Mid-360 & 64.29 & 24.86\\
    Early Fusion & Early Fusion & \textbf{96.28} & \textbf{61.41}\\
    Early Fusion & Livox Mid-360 & 66.69 & 27.24\\
    Early Fusion & Velodyne VLP-16 & 85.85 & 44.97\\
    \hline
    \end{tabular}
    \caption{Recall@1\% and Recall@1 results of the cross-sensor evaluation.}
    \label{tab:cross_sensor}
\end{table}

The results demonstrate that our method performs well in cross-sensor scenarios, with only a slight decrease in performance compared to training and testing on the same sensor. This is an indicator that, while the difficulty of the cross-sensor scenario is significant, our fusion strategy allows the model to learn more robust features that can generalize across different sensors. Furthermore, the results also show that training on fused data and testing on each of the sensors separately also yields better performance than training and testing on the same sensor, which further demonstrates the effectiveness of our fusion strategy in improving the generalization ability of the model across different sensors.

\subsection{Impact of re-ranking}\label{subsec:re-ranking_ablation}
In order to prove the efectiveness of our proposed learned re-ranking approach, we compare the results of our method before and after applying the re-ranking strategy. We evaluate the impact of this re-ranking head in each isolated sensor scenario, as well as with our early fusion approach. We also evaluate the impact of different input descriptors for the re-ranking head, such as the query descriptor ($q$), the candidate descriptor ($d$), their element-wise product ($q*d$), and their element-wise difference ($q-d$). This is because the product and difference of these descriptors can capture different aspects of the similarity and dissimilarity between queries and candidates. In order to prove the effectiveness of our approach, we provide results with the three phenological cycle packages: \textit{early growth}, \textit{flowering} and \textit{pre-harvest}, as was commented in the introduction of Section \ref{sec:ablation_study}. The results are presented in Tables \ref{tab:reranking_ablation_early}, \ref{tab:reranking_ablation_mid}, and \ref{tab:reranking_ablation_late}, respectively. 

Additionally, we evaluate the performance of the re-ranker on the BLT dataset \cite{polvara2024bacchus}. The BLT dataset only contains data from a single LiDAR 3D sensor, which is an Ouster OS1-16. By evaluating our re-ranking approach on this different dataset, we can demonstrate its generalization ability and its potential applicability to other agricultural environments beyond the one captured in the TEMPO-VINE dataset. To create a testbed similar to the \textit{flowering} phase using the TEMPO-VINE dataset, we selected campaign data from June (early, middle, and late), comprising four trajectories captured within the same month. The results of this evaluation are presented in Table \ref{tab:reranking_blt}.

\begin{table}[h]
    \centering
    \begin{tabular}{c|c|c|cc}
    \hline
    \textbf{Sensor Data} & \textbf{Re-ranking} & \textbf{Input desc.} & \textbf{Recall@1\%} & \textbf{Recall@1}\\
    \hline
    Livox Mid-360 & No & - & 87.25 & 44.81\\
    Livox Mid-360 & Yes & $q, d$ & \textbf{93.40} & 76.01\\
    Livox Mid-360 & Yes & $q, d, q*d$ & 93.36 & \textbf{76.16}\\
    Livox Mid-360 & Yes & $q, d, q-d$ & 93.27 & 75.17\\
    Livox Mid-360 & Yes & $q, d, q-d, q*d$ & 93.17 & 74.48\\
    \hline
    Velodyne VLP-16 & No & - & 98.01 & 75.07\\
    Velodyne VLP-16 & Yes & $q, d$ & 99.73 & 94.79\\
    Velodyne VLP-16 & Yes & $q, d, q*d$ & \textbf{99.79} & \textbf{96.51}\\
    Velodyne VLP-16 & Yes & $q, d, q-d$ & 99.75 & 93.95\\
    Velodyne VLP-16 & Yes & $q, d, q-d, q*d$ & 99.77 & 94.25\\
    \hline
    Early Fusion & No & - & 99.94 & 83.60\\
    Early Fusion & Yes & $q, d$ & \textbf{100} & 96.40\\
    Early Fusion & Yes & $q, d, q*d$ & 99.98 & 96.25\\
    Early Fusion & Yes & $q, d, q-d$ & \textbf{100} & \textbf{96.85}\\
    Early Fusion & Yes & $q, d, q-d, q*d$ & \textbf{100} & 96.62\\ 
    \hline
    \end{tabular}
    \caption{Recall@1\% and Recall@1 results comparison before and after our proposed re-ranking strategy for each of the sensor configurations in the \textit{early growth} phenological state.}
    \label{tab:reranking_ablation_early}
\end{table}

\begin{table}[h]
    \centering
    \begin{tabular}{c|c|c|cc}
    \hline
    \textbf{Sensor Data} & \textbf{Re-ranking} & \textbf{Input desc.} & \textbf{Recall@1\%} & \textbf{Recall@1}\\
    \hline
    Livox Mid-360 & No & - & 62.79 & 26.85\\
    Livox Mid-360 & Yes & $q, d$ & 72.06 & 53.23\\
    Livox Mid-360 & Yes & $q, d, q*d$ & \textbf{72.33} & \textbf{53.86}\\
    Livox Mid-360 & Yes & $q, d, q-d$ & 72.17 & 53.69\\
    Livox Mid-360 & Yes & $q, d, q-d, q*d$ & 72.08 & 53.04\\
    \hline
    Velodyne VLP-16 & No & - & 88.38 & 52.80\\
    Velodyne VLP-16 & Yes & $q, d$ & 94.70 & 81.94\\
    Velodyne VLP-16 & Yes & $q, d, q*d$ & \textbf{94.80} & \textbf{83.29}\\
    Velodyne VLP-16 & Yes & $q, d, q-d$ & 94.76 & 80.85\\
    Velodyne VLP-16 & Yes & $q, d, q-d, q*d$ & \textbf{94.80} & 82.00\\
    \hline
    Early Fusion & No & - & 96.28 & 61.41\\
    Early Fusion & Yes & $q, d$ & 98.40 & \textbf{89.90}\\
    Early Fusion & Yes & $q, d, q*d$ & 98.41 & 87.69\\
    Early Fusion & Yes & $q, d, q-d$ & \textbf{98.47} & 87.98\\
    Early Fusion & Yes & $q, d, q-d, q*d$ & 98.30 & 87.43\\ 
    \hline
    \end{tabular}
    \caption{Recall@1\% and Recall@1 results comparison before and after our proposed re-ranking strategy for each of the sensor configurations in the \textit{flowering} phenological state.}
    \label{tab:reranking_ablation_mid}
\end{table}

\begin{table}[h]
    \centering
    \begin{tabular}{c|c|c|cc}
    \hline
    \textbf{Sensor Data} & \textbf{Re-ranking} & \textbf{Input desc.} & \textbf{Recall@1\%} & \textbf{Recall@1}\\
    \hline
    Livox Mid-360 & No & - & 86.89 & 50.04\\
    Livox Mid-360 & Yes & $q, d$ & 81.66 & \textbf{63.86}\\
    Livox Mid-360 & Yes & $q, d, q*d$ & \textbf{81.88} & 63.17\\
    Livox Mid-360 & Yes & $q, d, q-d$ & 81.49 & 61.18\\
    Livox Mid-360 & Yes & $q, d, q-d, q*d$ & 81.62 & 59.91\\
    \hline
    Velodyne VLP-16 & No & - & 97.38 & 64.45\\
    Velodyne VLP-16 & Yes & $q, d$ & 99.38 & \textbf{92.12}\\
    Velodyne VLP-16 & Yes & $q, d, q*d$ & \textbf{99.50} & 91.99\\
    Velodyne VLP-16 & Yes & $q, d, q-d$ & 99.42 & 90.68\\
    Velodyne VLP-16 & Yes & $q, d, q-d, q*d$ & 99.31 & 89.30\\
    \hline
    Early Fusion & No & - & 98.88 & 73.33\\
    Early Fusion & Yes & $q, d$ & 99.50 & \textbf{94.03}\\
    Early Fusion & Yes & $q, d, q*d$ & \textbf{99.63} & \textbf{94.03}\\
    Early Fusion & Yes & $q, d, q-d$ & 99.48 & 90.20\\
    Early Fusion & Yes & $q, d, q-d, q*d$ & 99.48 & 90.59\\ 
    \hline
    \end{tabular}
    \caption{Recall@1\% and Recall@1 results comparison before and after our proposed re-ranking strategy for each of the sensor configurations in the \textit{pre-harvest} phenological state.}
    \label{tab:reranking_ablation_late}
\end{table}

\begin{table}[h]
    \centering
    \begin{tabular}{c|c|cc}
    \hline
    \textbf{Re-ranking} & \textbf{Input desc.} & \textbf{Recall@1\%} & \textbf{Recall@1}\\
    \hline
    No & - & 76.82 & 40.46\\
    Yes & $q, d$ & \textbf{87.33} & 86.02\\ 
    Yes & $q, d, q*d$ & \textbf{87.33} & 85.82\\ 
    Yes & $q, d, q-d$ & \textbf{87.33} & 85.82\\ 
    Yes & $q, d, q-d, q*d$ & \textbf{87.33} & \textbf{86.19}\\ 
    \hline
    \end{tabular}
    \caption{Recall@1\% and Recall@1 results comparison before and after our proposed re-ranking strategy on the BLT dataset. The point clouds were collected with a single Ouster OS1-16 LiDAR sensor.}
    \label{tab:reranking_blt}
\end{table}

The results show that the learned re-ranking approach significantly improves the performance of our method in all sensor configurations. Regarding the TEMPO-VINE dataset, the improvement is particularly notable in the fused data scenario, where we see an increase of over 28\% in Recall@1 when applying the re-ranking strategy. This demonstrates the importance of re-ranking strategies in refining the results of initial retrieval stages, especially in unstructured environments where the presence of occlusions and noise can lead to a high number of false positives. In the case of the BLT dataset (Table \ref{tab:reranking_blt}), we also see a significant improvement in performance when applying the re-ranking strategy, with an increase of over 45\% in Recall@1. Moreover, the difference between Recall@1\% and Recall@1 is much smaller in the re-ranked results. This further demonstrates the generalization ability of our proposed re-ranking approach across different datasets and environments.

Regarding the selection of input descriptors for the re-ranking head, the different tables reveal that it is challenging to identify a consistent pattern indicating which combination of descriptors provides the best performance. However, we can see that the combination of query and candidate descriptors ($q, d$) is crucial for the performance of the re-ranking head, as it consistently provides a significant improvement compared to not applying re-ranking. The addition of the product and difference descriptors ($q*d$ and $q-d$) can provide further improvements in some cases, but it is not consistent across all scenarios. Therefore, we can conclude that while the combination of query and candidate descriptors is essential for the re-ranking head, the inclusion of product and difference descriptors may provide additional benefits in certain cases, but it is not a guaranteed improvement in all scenarios.

In order to decide the final configuration of input descriptors for the re-ranking head, we also analyze the computational performance of each configuration, as will be explained in the next section.

\subsubsection{Computational analysis of the re-ranking head}\label{subsubsec:ablation_reranking_computational_analysis}
With the aim of choosing a re-ranking head that provides a good balance between performance and computational efficiency, we analyze the computational performance of our proposed re-ranking head with different input descriptors. We show the total number of parameters, Floating-point Operations (FLOPs), and Multiply-Accumulate operations (MACs) for each configuration. The MACs are obtained through dividing the number of FLOPs by two \cite{dong2022hardware}. The number of FLOPs was obtained using the python pvcore library. The evaluation was performed on an Nvidia A30 GPU. 

The values related to these metrics are presented in Table \ref{tab:reranking_computational_analysis}. Based on this table, we can see that the configuration with only the query and candidate descriptors ($q, d$) has the lowest computational cost, with 115,073 parameters and 114,752 FLOPs. The addition of the product and difference descriptors ($q*d$ and $q-d$) increases the computational cost significantly, with the configuration that includes both product and difference descriptors ($q, d, q-d, q*d$) having the highest computational cost with 213,377 parameters and 213,056 FLOPs. Given that the performance improvement from adding the product and difference descriptors is not consistent across all scenarios, and considering the significant increase in computational cost, we decide to use the configuration with only the query and candidate descriptors ($q, d$) for our re-ranking head. This configuration provides a good balance between performance improvement and computational efficiency, making it suitable for real-time applications in unstructured environments.

\begin{table}[h]
    \centering
    \begin{tabular}{c|ccc}
    \hline
    \textbf{Input desc.} & \textbf{Params.} & \textbf{FLOPs} & \textbf{MACs}\\
    \hline
    $q, d$ & 115,073 & 114,752 & 57,376\\
    $q, d, q*d$ & 164,225 & 163,904 & 81,952\\
    $q, d, q-d$ & 164,225 & 163,904 & 81,952\\
    $q, d, q-d, q*d$ & 213,377 & 213,056 & 106,528\\ 
    \hline
    \end{tabular}
    \caption{Total number of parameters, FLOPs, and MACs of the re-ranking head for each configuration of input descriptors.}
    \label{tab:reranking_computational_analysis}
\end{table}

\newpage

\subsection{Cross-season evaluation}\label{subsec:ablation_cross_season}
In this section, we evaluate the performance of our MinkUNeXt-VINE++ in a cross-season scenario. We group the campaigns into the three different phenological stages mentioned previously: \textit{early growth}, \textit{flowering}, and \textit{pre-harvest}. We train our MinkUNeXt-VINE++ using data from one phenological stage and test it using data from a different phenological stage to show the cross-season generalization capability. 

\begin{figure}[h]
    \centering
    \includegraphics[width=\textwidth]{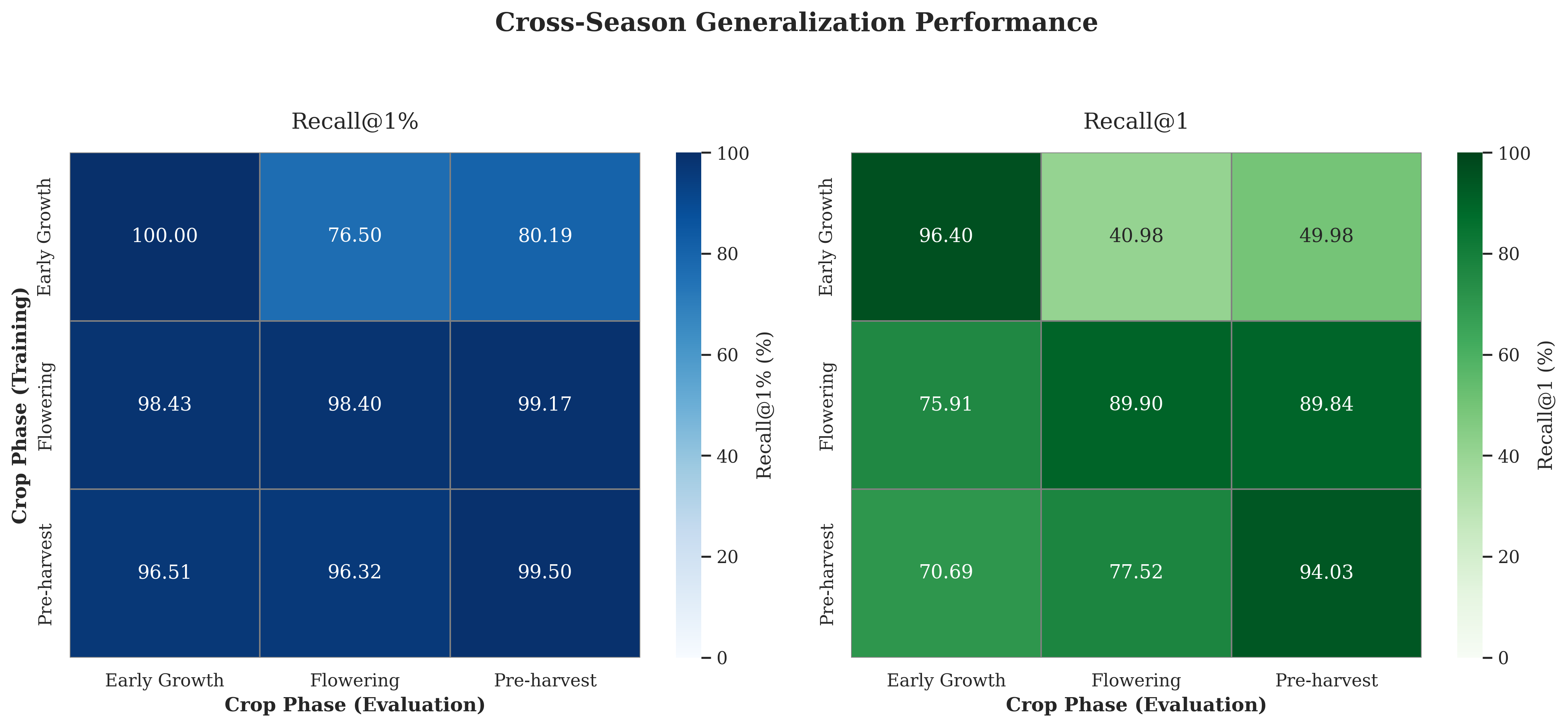}
    \caption{Recall@1\% and Recall@1 heatmap results of MinkUNeXt-VINE++ in the cross-season evaluation based on phenological stages.}
    \label{fig:cross_season_phenological}
\end{figure}

Figure \ref{fig:cross_season_phenological} shows the Recall@1 heatmap results of MinkUNeXt-VINE++ in the cross-season evaluation based on phenological stages. 

The heatmaps illustrate that, while there is a decline in performance when evaluating on a different phenological stage than the one used for training, our method maintains a satisfactory performance in the majority of the cross-season scenarios. This demonstrates that our proposed contributions, especially the fusion strategy, allow the model to learn more robust and generalizable features that can adapt to different environmental conditions and phenological stages. The best performance is observed when training on the \textit{flowering} stage and testing on the \textit{pre-harvest} stage, with a Recall@1 of 89.84\%. This indicates that the features learned during the \textit{flowering} stage are particularly effective for recognizing places in the \textit{pre-harvest} stage, which may be due to the fact that both stages share similar visual characteristics in terms of vegetation density and structure. The performance drop is more significant when training on the \textit{early growth} stage and testing on the \textit{flowering} and the \textit{pre-harvest} stages, with a Recall@1 of 40.98\% and 49.98\%, respectively. This is likely due to the significant changes in vegetation structure and density between these stages, which can make it more challenging for the model to generalize. However, even in this challenging scenario, our method still achieves a reasonable performance, specially in the Recall@1\% metric, demonstrating adaptability to different phenological stages.

\section{Results and discussion}\label{sec:discussion}
In this section, we present the results of our combined contributions (fusion and re-ranking) and compare them with the results from other SOTA LPR solutions. We also discuss the implications of our findings and the potential impact of MinkUNeXt-VINE++ on the field of LiDAR-based place recognition in unstructured environments.

Table \ref{tab:final_results} presents the final evaluation of MinkUNeXt-VINE++ against state-of-the-art LPR methods, including PointNetVLAD \cite{uy2018pointnetvlad}, MinkLoc3Dv2 \cite{komorowski2022improving}, MinkUNeXt \cite{cabrera2025minkunext}, and our base method MinkUNeXt-VINE \cite{vilella2026low}. To ensure a fair representation of each method's intended use case, the baseline models are evaluated under their standard single-sensor configurations (Velodyne VLP-16 and Livox), whereas MinkUNeXt-VINE++ is evaluated using its native early fusion architecture. While exhaustive cross-configurations (evaluating the baselines with fused data and our method with single sensors) are thoroughly analyzed in our ablation study (Section \ref{subsec:ablation_fusion}), this table highlights the peak performance of each approach in its optimal operational setup.

\begin{landscape}
    \begin{table}[h]
        \centering
        \begin{tabular}{|c| >{\centering\arraybackslash}m{2cm}| >{\centering\arraybackslash}m{2cm}| >{\centering\arraybackslash}m{2cm}| >{\centering\arraybackslash}m{2cm}| >{\centering\arraybackslash}m{2cm}| >{\centering\arraybackslash}m{2cm}|}
        \hline
        \multirow{2}{*}{\textbf{Method}} & 
        \multicolumn{2}{c|}{\textbf{TEMPO-VINE (VELO)}} & 
        \multicolumn{2}{c|}{\textbf{TEMPO-VINE (Livox)}} & 
        \multicolumn{2}{c|}{\textbf{Early-Fusion}} \\ 
        \cline{2-7}
        & Recall@1\% & Recall@1 & Recall@1\% & Recall@1 & Recall@1\% & Recall@1 \\
        \hline
        PointNetVLAD \cite{uy2018pointnetvlad} & 30.55 & 5.54 & 25.15 & 4.23 & - & -\\
        MinkLoc3Dv2 \cite{komorowski2022improving} & 19.09 & 2.53 & 18.64 & 2.33 & - & -\\
        MinkUNeXt \cite{cabrera2025minkunext} & 20.77 & 6.43 & 16.62 & 3.08 & - & -\\
        MinkUNeXt-VINE \cite{vilella2026low} & 94.57 & 59.51 & 73.58 & 31.85 & - & -\\
        MinkUNeXt-VINE++ & - & - & - & - & \textbf{98.14} & \textbf{89.32}\\
        \hline
        \end{tabular}
        \caption{Final results of our method compared to other SOTA LPR methods on the TEMPO-VINE dataset.}
        \label{tab:final_results}
    \end{table}
\end{landscape}

As shown in Table \ref{tab:final_results}, standard general-purpose LPR methods struggle significantly when deployed in agricultural vineyard environments. PointNetVLAD \cite{uy2018pointnetvlad}, MinkUNeXt \cite{cabrera2025minkunext}, and MinkLoc3Dv2 \cite{komorowski2022improving} exhibit severely degraded performance across both individual sensor configurations. For instance, MinkLoc3Dv2 achieves a meager Recall@1\% of $30.53\%$ on Velodyne data and $25.15\%$ on Livox data, while its Recall@1 drops to nearly 5\% for both sensors. Although MinkUNeXt \cite{cabrera2025minkunext} provides a slight improvement, yielding a Recall@1\% of 20.77 on Velodyne, its top-1 retrieval accuracy remains low at 6.43. This widespread performance drop highlights the extreme challenges posed by agricultural environments. The integration of domain-specific refinements in MinkUNeXt-VINE \cite{vilella2026low} yields a substantial performance leap. When operating solely on Velodyne data, MinkUNeXt-VINE achieves an outstanding Recall@1\% of $94.57\%$ and a Recall@1 of $59.51\%$. A similar upward trend is visible on the Livox data, where it reaches $73.58\%$ (Recall@1\%) and $31.85\%$ (Recall@1). However, it is evident that, while the Recall@1\% metric indicates a strong ability to retrieve relevant candidates, the strict Recall@1 metric reveals that the model still struggles to identify the single best match in many cases.

Our proposed method, MinkUNeXt-VINE++, completely outperforms all previous approaches and single-sensor baselines, establishing a new state-of-the-art on the TEMPO-VINE dataset. By leveraging our dual contribution framework, MinkUNeXt-VINE++ achieves an enhanced Recall@1\% of $98.14\%$ and a Recall@1 of $89.32\%$, comparable to the results that are usually achieved under urban settings. Compared to the best-performing single-sensor baseline (MinkUNeXt-VINE on Velodyne), our method delivers an improvement of $+4\%$ in Recall@1\% and a remarkable $+30\%$ in Recall@1. This notable surge in performance for the Recall@1 metric validates the synergy between our two main technical contributions, the early fusion strategy and the learned re-ranking approach, demonstrating their effectiveness in addressing the unique challenges of place recognition in unstructured agricultural environments.

\section{Conclusions}\label{sec:conclusions}
In this paper, we presented MinkUNeXt-VINE++, which comprises two main contributions to enhance the performance of LiDAR-based place recognition in unstructured environments: a heterogeneous LiDAR early fusion strategy and a learned re-ranking approach. Our fusion strategy leverages the strengths of both Livox Mid-360 and Velodyne VLP-16 sensors, providing a more comprehensive representation of the environment. The learned re-ranking approach enhances the final ranking of candidate places, particularly in repetitive environments such as vineyards. We evaluated our method on the TEMPO-VINE dataset, as it includes heterogeneous LiDAR data, demonstrating significant improvements in place recognition performance compared to single-sensor approaches and SOTA methods. Our fusion method achieved a 20\% improvement in Recall@1 compared to single-sensor approaches, highlighting the effectiveness of our proposed contributions. In what regards the learned re-ranking approach, it presented a 28\% improvement in the Recall@1 metric with respect to not including it along with the early fusion approach. Furthermore, the re-ranking contribution also showed a significant improvement of 46\% of Recall@1 in the BLT dataset, which demonstrates its generalization ability across different datasets and environments. All the experiments in this paper were performed following a long-term focus, evaluating the performance of our method in cross-season scenarios to demonstrate its robustness and adaptability to different crop growth stages.

As future work, we plan to explore a more scalable fusion strategy that can be applied to a wider range of sensors and environments. Additionally, we aim to extend the application of our method to other environments, such as urban settings, to further validate its effectiveness and generalization capabilities. Regarding the re-ranking head, our next objective is to create an extensive benchmark of the effects of different input descriptors and seasonal changes on correct reordering in agricultural environments. Finally, our goal is to deploy our method on an autonomous system operating in a real-world vineyard environment to demonstrate its practical applicability and robustness.

\section*{Author statement}
We would like to make the following declarations: All authors confirm that this work is original and has not been published elsewhere, nor is it currently under consideration for publication elsewhere.

We confirm that the manuscript has been read and approved by all named authors and that there are no other persons who satisfy the criteria for authorship but are not listed. We further confirm that the order of authors listed in the manuscript has been approved by all of us.

We understand that the Corresponding Author is the sole contact for the Editorial process. He is responsible for communicating with the other authors about progress, submissions of revisions and final approval of proofs.

\section*{CRediT authorship contribution statement}
\textbf{Judith Vilella-Cantos:} Writing – review \& editing, Writing – original draft, Software, Methodology, Data curation, Conceptualization, Investigation. \textbf{Juan José Cabrera:} Software, Formal analysis, Conceptualization, Investigation. \textbf{Mónica Ballesta:} Supervision, Formal analysis, Writing – review \& editing, Project administration, Funding acquisition. \textbf{David Valiente:} Supervision, Formal analysis, Writing – review \& editing, Project administration, Funding acquisition. \textbf{Luis Payá:} Resources, Conceptualization, Project administration, Funding acquisition. 

\section*{Declaration of competing interest}
The authors declare that they have no known competing financial interests or personal relationships that could have appeared to influence the work reported in this paper.

\section*{Acknowledgements}
This research work is part of the project PID2023-149575OB-I00 funded by MICIU/AEI/10.13039/501100011033 and by FEDER, UE. It is also part of the project CIPROM/2024/8, funded by Generalitat Valenciana, Conselleria de Educación, Cultura, Universidades y Empleo (program PROMETEO 2025).

\section*{Data availability}
The data of our model is available for reproduction on a public Github repository. This data can be accessed through the following link: https://git\-hub.com/JudithV/MinkUNeXt-VINE\_plusplus.

\bibliographystyle{elsarticle-num}
\bibliography{bibliography}

\end{document}